\documentclass[letterpaper, 10 pt, conference]{ieeeconf}  
\IEEEoverridecommandlockouts                              
\usepackage{cite}
\usepackage{amsmath,amssymb,amsfonts}
\usepackage{algorithmic}
\usepackage{algorithm}
\usepackage{graphicx}
\usepackage{textcomp}
\usepackage{xcolor}
\usepackage[acronyms]{glossaries}
\usepackage{mathtools}
\usepackage{regexpatch}
\usepackage{svg}
\usepackage{enumerate}
\usepackage{flushend} 
\usepackage{tikz-network}
\usepackage{caption}
\usepackage{subcaption}
\usepackage{comment}
\usepackage{booktabs}
\usepackage{multirow}

\def\BibTeX{{\rm B\kern-.05em{\sc i\kern-.025em b}\kern-.08em
    T\kern-.1667em\lower.7ex\hbox{E}\kern-.125emX}}

\def\ie{\textit{i.e.},}
\def\eg{\textit{e.g.},}

\newcommand\numberthis{\addtocounter{equation}{1}\tag{\theequation}}

\newacronym{mcl}{MCL}{Monte-Carlo Localization}
\newacronym{ndt-mcl}{NDT-MCL}{Normal Distributions Transform Monte-Carlo Localization}
\newacronym[plural=NDT-OMs,firstplural=Normal Distributions Occupancy Maps (NDT-OM)]{ndt-om}{NDT-OM}{Normal Distributions Occupancy Map}
\newacronym{ndt-se}{NDT-SE}{Semantically Enhanced Normal Distributions Transform}
\newacronym{d2d}{D2D}{Distribution-to-Distribution}
\newacronym{ins}{INS}{Inertial Navigation System}
\newacronym{ned}{NED}{North-East-Down}
\newacronym{utm}{UTM}{Universal Transverse Mercator}
\newacronym{rmse}{RMSE}{Root Mean Squared Error}
\newacronym{ate}{ATE}{Absolute Trajectory Error}
\newacronym{rpe}{RPE}{Relative Pose Error}
\newacronym{hmm}{HMM}{Hidden Markov Model}
\newacronym{av}{AV}{autonomous vehicle}
\newacronym{ros}{ROS}{Robot Operating System}
\newacronym{gps}{GPS}{Global Positioning System}
\newacronym{sd-ndt-om}{SD-NDT-OM}{Semantic-Dynamic Normal Distributions Occupancy Map}
\newacronym{slam}{SLAM}{Simultaneous Localization and Mapping}

\def\figvspace{\vspace{1em}}
\def\figvspacelast{\vspace{0.5em}}
\def\tabcapspace{\vspace{0.6em}}

\title{\LARGE \bf
Localization Under Consistent Assumptions Over Dynamics
}

\author{Matti Pekkanen, Francesco Verdoja, and Ville Kyrki 
\thanks{This work was supported by Business Finland, decision 9249/31/2021. We gratefully acknowledge the support of NVIDIA Corporation with the donation of the Titan Xp GPU used for this research.}
\thanks{M. Pekkanen, F. Verdoja and V. Kyrki are with School of Electrical Engineering,
        Aalto University, Espoo, Finland.
        {\tt\small \{firstname.lastname\}@aalto.fi}}%
}
\begin{document}
\bstctlcite{IEEEexample:BSTcontrol}
\maketitle
\thispagestyle{empty}
\pagestyle{empty}

\begin{abstract}
    Accurate maps are a prerequisite for virtually all mobile robot tasks. Most state-of-the-art maps assume a static world; therefore, dynamic objects are filtered out of the measurements. However, this division ignores movable but non-moving---\ie{} semi-static---objects, which are usually recorded in the map and treated as static objects, violating the static world assumption and causing errors in the localization. This paper presents a method for consistently modeling moving and movable objects to match the map and measurements. This reduces the error resulting from inconsistent categorization and treatment of non-static measurements. A semantic segmentation network is used to categorize the measurements into static and semi-static classes, and a background subtraction filter is used to remove dynamic measurements. Finally, we show that consistent assumptions over dynamics improve localization accuracy when compared against a state-of-the-art baseline solution using real-world data from the Oxford Radar RobotCar data set.
\end{abstract}

\section{Introduction}
\label{sec:intro}

Mapping is a central functionality of mobile robot systems since an accurate representation of the environment, \ie{} a map, is a prerequisite for many crucial functionalities, such as localization and path planning.

Most existing mapping methods assume that the mapped environment does not change until the map is used for localization. This is usually referred to as the \emph{static world assumption}. The assumption is made for simplicity, even if it does not entirely hold. Violations of the assumption, however, may result in errors in localization.

\begin{figure}[t]
\centering
\includegraphics[width=0.48\textwidth]{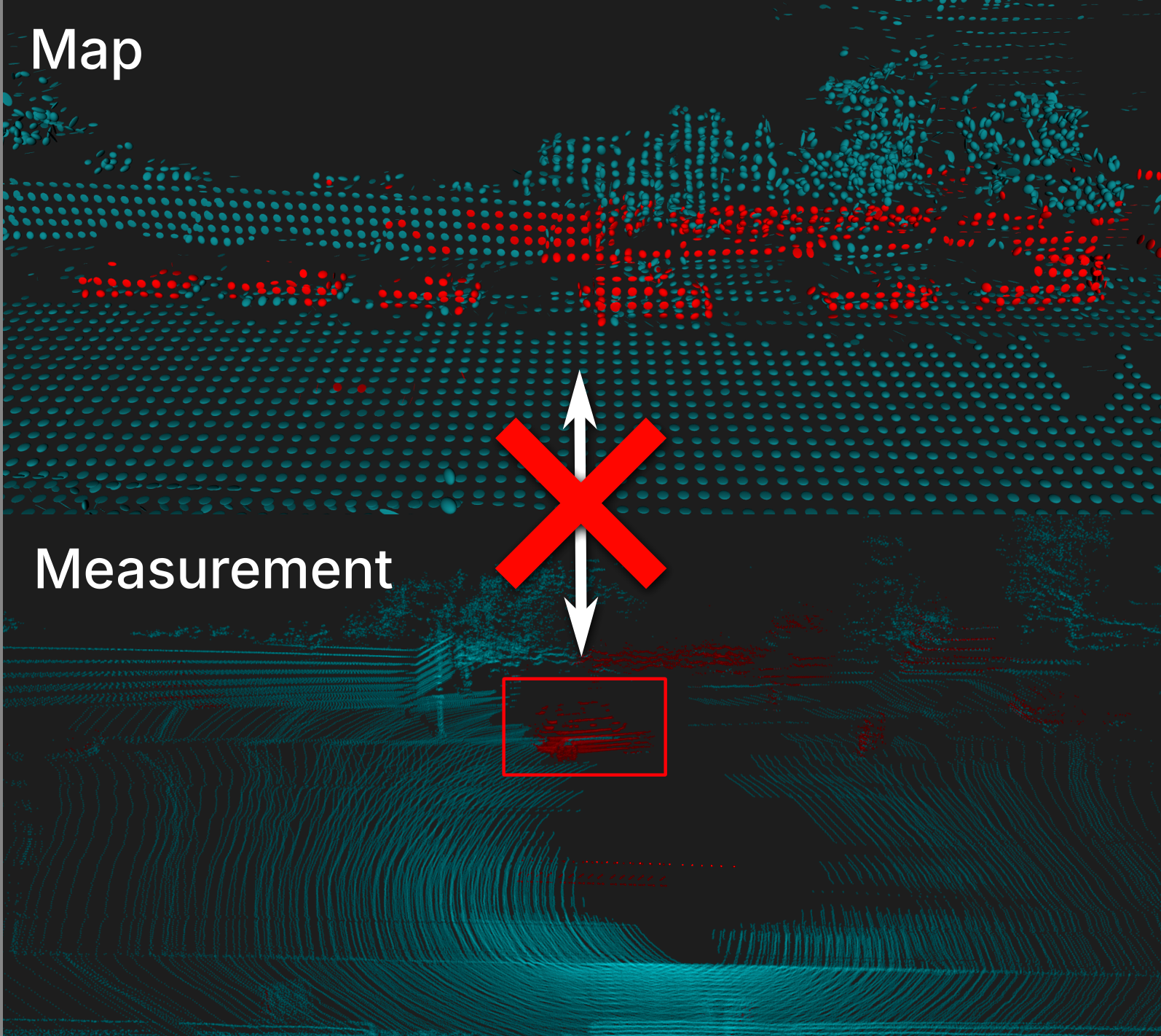}
\caption{Treating semi-static objects as static ones violates the static world assumption and causes mismatches between measurements and a map. In the figure, the map contains parked cars that have since moved away. When the observer returns, a new parked car is detected offset from the ones on the map. This causes matching errors, especially when the difference in poses is small or other features in the direction of the error are lacking or sparse.}
\label{fig:mismatch}
\figvspace{}
\end{figure}

For example, the map might contain parked cars, which would be considered equally reliable landmarks compared to non-movable, \ie{} \emph{static}, objects such as buildings. If, during localization, another car was observed in a different pose than the car on the map that has since left, the potential incorrect match may cause a localization error. This phenomenon is illustrated in Figure \ref{fig:mismatch}.

To address this problem, many methods for removing moving, \ie{} \emph{dynamic}, objects from the measurements have been proposed \cite{fox_position_1998, hahnel_map_2003, wolf_mobile_2005, kummerle_navigation_2013, saarinen_ndtom_2013, chen_suma_2019}, and it continues to be the most common approach in the state-of-the-art localization and mapping methods. This dichotomy between moving and non-moving objects ignores objects that are \emph{movable} while not currently moving. In this work, we call such objects \emph{semi-static objects} and assume that the environment consists of objects from these three \emph{dynamic classes}: static, semi-static, and dynamic.

With the increased performance of semantic segmentation networks, detecting semi-static objects directly from laser measurements is possible. However, proper representation of the dynamic classes in maps is still rare, and semi-static objects are usually treated as static, violating the static world assumption.

In this work, we propose a better way: by distinguishing between the properties of movability and motion, we can properly model the dynamic, semi-static, and static parts of the environment. The consistent application of this distinction complies not only with the static world assumption but also all our \emph{assumptions over dynamics}. Using real-world data from real traffic scenarios gathered over nine days, we show that localization under consistent assumptions over dynamics increases localization accuracy.

We can partition the measurements into dynamic classes by using semantic segmentation of laser point clouds, background subtraction, and clustering-based dynamic object filtering. Using these filters to be consistent in the assumptions over dynamics, we create an \gls{ndt-om} \cite{saarinen_ndtom_2013} containing only static measurements. For comparison, we use the state-of-the-art baseline \gls{ndt-om}, which does not discriminate between semi-static and static measurements and, therefore, violates the static world assumption.

Similarly, we use the aforementioned filters to demonstrate four localization methods based on \gls{ndt-mcl} \cite{saarinen_mcl_2013}, each using measurements of different dynamic classes in the localization. Subsequently, we show that localization accuracy is best when we match the measurements with the maps under consistent assumptions over dynamics.

The main contributions of this paper are:
\begin{enumerate}[i)]
    \item We propose a localization method using semantic segmentation and dynamic filtering to remove non-static measurements from the input measurements of the localization.
    \item We propose a mapping method using semantic segmentation to remove non-static measurements to produce a map compliant with the static world assumption.
    \item We show with an empirical study consisting of 112 localization experiments that the localization accuracy of the baseline \gls{ndt-mcl} can be improved using the proposed mapping method to create a map consisting of only static measurements and the proposed localization method.
\end{enumerate}

\begin{figure*}[t]
\centering
\includegraphics[width=\textwidth]{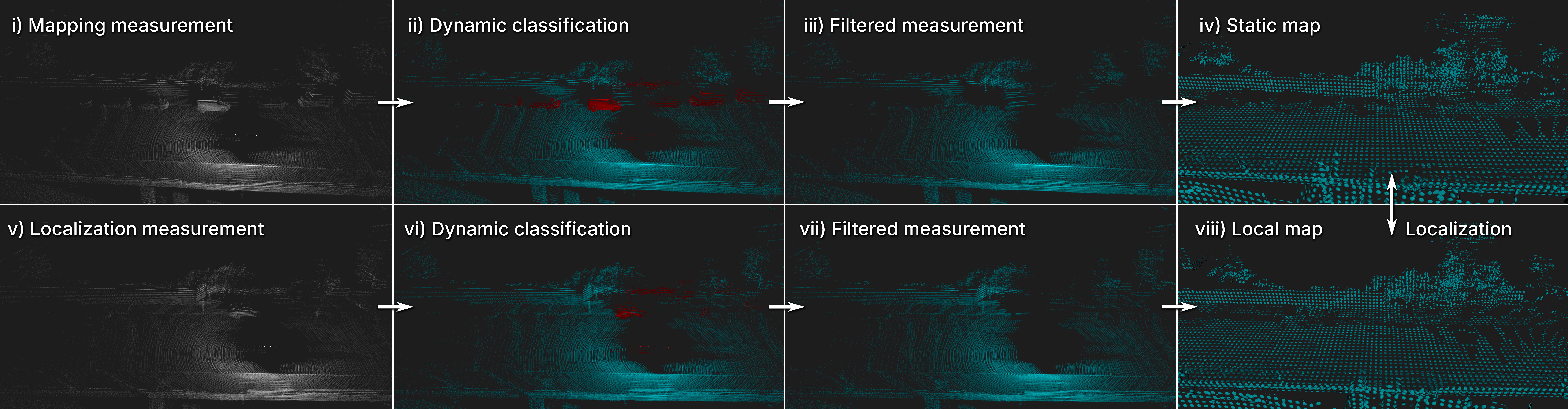}
\caption{Overview of the proposed method. The mapping process is depicted in the first row, and the localization process is depicted in the second row. First, in i) and v), the dynamic classes for each measurement point are estimated. Then, the dynamic and semi-static objects, shown in red, are removed from the measurements used for building a map ii) and localizing vi). This results in measurements where only the reliable static measurements are used, shown in blue in iii) and vii). This allows the localization using consistent assumptions over dynamics between the map iv) and the measurement viii). This is in contrast to the state-of-the-art localization methods, where measurements are directly compared to the map, both containing semi-static and dynamic objects, causing localization errors.}
\label{fig:method}
\figvspace{}
\end{figure*}
\section{Related work}
\label{sec:related}

\subsection{Filtering dynamic objects}

The most commonly used map type in mobile robotics is the occupancy map \cite{elfes_sonar-based_1987}. Occupancy maps incorporate the static world assumption, as they do not model the dynamic properties of the content of the cells.

While dynamic objects appear on the maps, they are removed after the occupied space has been observed empty by the free space modeling of the inverse sensor model. While this approach is widely used, it has several problems. For the dynamic objects to be removed, the space must be perceived as empty, so at the end of the mapping sequence or when transitioning to a different area, it is likely that dynamic objects will remain on the map. Additionally, the map does not represent the dynamic properties of the environment, so the sensor model cannot take into account the dynamic properties of the observed object when updating the occupancy probabilities of the affected grid cells.

To alleviate these problems, methods to filter dynamic objects from the measurements have been proposed \cite{fox_position_1998, hahnel_map_2003, wolf_mobile_2005, kummerle_navigation_2013}. However, none of these approaches distinguishes between static and semi-static objects; subsequently, they all leave the semi-static objects in the map. In this work, we treat all objects consistently according to their dynamic classes, including semi-static ones.

\subsection{Representation of semi-static objects}

Several methods have been proposed to address the issue of semi-static objects being treated as static. Semi-static objects have been represented as separate temporary maps \cite{meyer-delius_temporary_2010} with a given static map. Instead of modeling the dynamics explicitly, \cite{meyer-delius_temporary_2010} stores multiple maps from different times as static snapshots of the different states of the environment and selects the map that best explains the current measurement. While this approach is not intended to create a consistent representation of the environment but rather to facilitate localization, the idea extends to similar methods that jointly localize the robot and estimate the state of the environment. This has been demonstrated multiple times with, \eg{} a door \cite{petrovskaya_probabilistic_nodate, schulz_probabilistic_2001}.

A step forward in representing the dynamic nature of the environment is to model the environment as an \gls{hmm} \cite{tipaldi_lifelong_2013, meyer-delius_occupancy_2021}. While an \gls{hmm} explicitly models the belief of occupancy and the transition probabilities of the environment, which can be used to improve localization accuracy, it makes no distinction between dynamic or static cells, unlike this work.

Furthermore, the static world assumption is ingrained in the Markov assumptions of independence of odometry and observations. These assumptions have been relaxed by partitioning the localization experiment into internally Markovian episodes, but as a whole, the experiment is not \cite{biswas_episodic_2014}. Instead, in this work, we aim to maintain a single, consistent environment representation.

While static objects are considered not movable, semi-static objects are likely to move during the lifetime of a map. Therefore, the probability of any object remaining stationary reduces over time. This degree of staticness can be modeled explicitly as the decaying probability of the persistence of a feature \cite{rosen_towards_2016}. Features are more naturally linked to object instances that can be ascribed with a notion of staticness, whereas we directly model the dynamic properties of the entire spatial environment. 

Finally, while parking lots have been the dominant testing environment for modeling semi-static objects \cite{meyer-delius_occupancy_2021, zhu_lifelong_2021, adkins_probabilistic_2022}, we extend the study on the effect of semi-static objects to more complex real-world urban scenarios.

\subsection{Using semantic segmentation}

While dynamic objects can be detected directly from the differences between subsequent measurements, semi-static objects can not. This problem can be solved using semantic segmentation to label objects with a semantic class. Using prior human experience, a certain set of labels can be categorized as movable, while the complement of the set is the unmovable objects. A prevalent type of movable semi-static object is a parked car.

A method for augmenting an NDT map with semantic information is proposed in \cite{zaganidis_semantically_2019}, where a separate \gls{ndt-om} is created for each semantic label. In registration, the measurements are partitioned according to the labels, and the measurements are registered against the map with the same labels as the measurements. Unlike this work, that method trusts each semantic class equally without addressing whether the object is static, semi-static, or dynamic.

Semantic segmentation has been leveraged to filter dynamic objects from the map \cite{chen_suma_2019}. In that work, all points belonging to movable classes were removed, whether they were moving or not. Instead, this work explicitly distinguishes semi-static and dynamic objects and models their dynamics accordingly.

To segment objects, a common solution is to use a combination of laser and camera \cite{zhu_lifelong_2021}. Images contain a richer amount of semantic information, which simplifies the segmentation task. The labels can be projected onto the laser point cloud after an image segmentation network is used to segment the camera image. However, the more desirable alternative, used in this work, is direct point-wise semantic labels for the laser. In combined laser and camera systems, the labels are constrained by the resolution and the field of view of the camera system, which can differ significantly from those of the laser system. A laser usually functions in dark and adverse weather conditions, whereas a camera does not.
\section{Problem statement}
\label{sec:problem}

The generic localization problem is defined as finding the posterior distribution of $p(x_{t} | z_{0:t}, u_{0:t}, m_{t})$, where $x_t$ is the state \ie{} the estimated pose at time $t$, $z_{0:t}$ the sequence of sets of measurements $z_{0:t} = \{z_0, ..., z_t\}$, where the set of measurements $z_t = \{z_t^0, ... z_t^n\}$ consists of individual measurements, $u_{0:t}$ is the set of control signals $u_{0:t} = \{u_0, ..., u_t\}$, and $m_t$ is the current state of the environment. 

In localization at time $t$, a map created previously at time $t_m$ is commonly used: $m_{t_m} \approx m_t, t_m \ll t$. However, this approximation holds only for the static parts of the environment. To solve the posterior through Bayes' theorem, the problem is finding a model of the measurement likelihood $p(z_{t} | x_t, m_{t_m})$, which would take into account that semi-static and dynamic parts of the environment might have moved. 
\section{Preliminaries}
\label{sec:preliminaries}

\subsection{Definitions}

To model the dynamics of objects, two properties of dynamics need to be considered: \textit{movability} (whether an object can move) and \textit{motion} (whether it is currently moving). The categorization of unmovable and movable objects depends on the context, \eg{} buildings can get demolished. However, we define unmovable objects as ones very unlikely to move during the lifetime of the map. We assume that movability depends on the semantic label of the object.

In terms of movability and motion, we consider that objects are partitioned into three dynamic classes:
\begin{itemize}
    \item Static $\mathcal{S}$: objects that are unmovable.
    \item Semi-static $\mathcal{E}$: objects that are movable but not in motion. 
    \item Dynamic $\mathcal{D}$: objects that are in motion.
\end{itemize}
Other similar definitions exist, such as \cite{meyer-delius_temporary_2010, morris_simultaneous_2014, zhu_lifelong_2021}.

We assume that movability is stationary over time; that is, unmovable objects cannot become movable, and vice versa. On the other hand, semi-static objects may start moving and become dynamic. So, the dynamic class can change, but the property of movability cannot. Additionally, we assume that the dynamic properties are distinct and must be estimated independently. Therefore, if an object is not in motion, its movability cannot be inferred from that fact alone. These assumptions are consistent with the real properties of objects. Therefore, we call these \emph{consistent assumptions over dynamics}.

\subsection{Semantic segmentation}

To estimate the dynamic class $d_z$ of a measurement, learning its semantic class is necessary. Let $\mathcal{L}$ be the set of all semantic labels. Let $\mathcal{L_D}$, $\mathcal{L_E}$ and $\mathcal{L_S}$ be the sets of all dynamic, semi-static, and static labels, respectively, such that the sets form a partition of $\mathcal{L}$.

Let $z_t$ be the set of all measurements from time $t$ with associated semantic labels. Let $z_t^d$ be the set of all measurements with label $l_z \in \mathcal{L_D}$, $z_t^e$ with label $l_z \in \mathcal{L_E}$ and $z_e^s$ with label $l_z \in \mathcal{L_S}$, such that these sets form a partition of $z_t$ according to the dynamic class.
\section{Methods}
\label{sec:methods}

We propose a method, shown in Figure \ref{fig:method}, to enable likelihood estimation with consistent assumptions over dynamics with any measurement model. The method may be used in localization, mapping, and \gls{slam}, and it consists of the following steps:
\begin{enumerate}
    \item At initialization, the set of selected dynamic classes $\delta_z \subseteq \{ \mathcal{S}, \mathcal{E}, \mathcal{D}\}$ is created.
    \item At each time $t$, a function $f_d(z_t^i)$ maps each measurement $z_t^i \in z_t$ to a dynamic class $d_z^i = f_d(z_t^i), d_z^i \in \{ \mathcal{S}, \mathcal{E}, \mathcal{D}\}$.
    \item Using the acquired dynamic classes, a subset of measurements $\Tilde{z}_t \subseteq z_t$ is selected such that it consists of only the measurements belonging to a set of selected dynamic classes $\delta_z$.
        \begin{align*}
        \Tilde{z}_t &= \{ z_t^i \in z_t : d_z^i \in \delta_z \} \numberthis
        \end{align*}
\end{enumerate}

When the method is used in map building, it yields a map, $\Tilde{m}$, consisting of only measurements of the selected dynamic classes for map building, $\delta_m$.

When the method is applied in localization, using the acquired subset of measurements $\Tilde{z}_t$, and the map $\Tilde{m}$, the original measurement model,
\begin{align*}
    &p(\Tilde{z}_t| x_t, \Tilde{m}), \numberthis
\end{align*}
comprises the given set of assumptions over dynamics, defined by $\delta_z$ and $\delta_m$.

This formulation has the benefit of leaving the definitions of the function $f_d(z)$, the map $m$, whether discrete or continuous, and the model $p(z_t|x_t,m)$ open for various implementations while enforcing constraints over dynamics. Our proposed implementation for the function $f_d(z)$ is discussed in Section \ref{sec:filters}, and the used map and measurement model in Section \ref{sec:map-creation}.

When the method is used with the selection $\delta_z = \delta_m = \{ \mathcal{S} \}$, the localization is consistent over assumptions over dynamics.
\section{Experiments}
\label{sec:experiments}

The two main questions we want to answer with the experiments are:
\begin{enumerate}
    \item Does the localization accuracy increase when the dynamic properties of the environment are better represented in the map content and/or the measurements?
    \item Does the localization accuracy decrease over time from map creation? Does this depend on the dynamic properties of the map content or the measurements?
\end{enumerate}

To answer these questions, we performed a series of experiments. We tested the proposed mapping method against the baseline \gls{ndt-om}. We created a map with each method on two sequences from the data set, for a total of four maps. Four localization methods were assessed using seven sequences for each map, totaling 112 localization experiments.

\subsection{Data set}

The Oxford Radar RobotCar data set \cite{RobotCarDatasetIJRR, RadarRobotCarDatasetICRA2020} was used in the experiments. This data set was selected as it consists of multiple traversals along the same route over multiple days, permitting the study of the effects of semi-static objects on localization accuracy, as the semi-static objects could move between the mapping and localization time. For this reason, the widely-used KITTI data set \cite{Geiger2013IJRR} could not be used, as each path in that data set is traversed in full only once.

The Oxford Radar RobotCar data set consists of 32 sequences where approximately the same route is traversed. The data set consists of data from seven different days over nine days. Nine sequences were selected from the data set: two for mapping and seven for localization. Maps were created from the first and last day of the data set, allowing the environment to change maximally between the sequences. For localization, the first sequence of each day was chosen unless the recording contained measurement failures, except for the first and last day, where the first sequence of the day was used for mapping and the second for localization.

\subsection{Sensor setup}

The Oxford Radar RobotCar has two Velodyne 32E lasers, of which the measurements from the left laser were used; the odometry was produced by NovAtel \gls{ins} system which consists of absolute position estimate in \gls{utm} coordinates, as well as linear velocity estimates and roll, pitch and yaw angles ($\varphi, \theta, \psi$) in \gls{ned} frame of reference.

Let $\prescript{W}{O}{T_0^1}$ be the transform from the odometry frame $O$ to the world frame $W$ at time $t = 0$ of the sequence 1. For practical purposes, all odometry measurements from all sequences were transformed such that $i$:th odometry measurement from sequence $s$, $\prescript{O}{}T_i^s$, was transformed
\begin{align*}
    \prescript{W}{}{\Tilde{T}_i^s} = \prescript{W}{O}{T_0^1} \cdot \prescript{O}{}{T_i^s},    \numberthis
\end{align*}

which sets $\prescript{W}{}{\Tilde{T}_0^1} = I$.

The data set contains the optimized $SE(2)$ solution for the NavTech CTS350-X radar, which was used as the ground truth. The ground truth solutions are relative to the starting pose, \ie{} $x_{t=0} = (0,0,0)$, so to enable comparison with the localization pose estimates, the ground truth was transformed to the world reference frame by minimizing the \gls{rmse} between the 2D translations of the transformed odometry and the ground truth.

\subsection{Semantic segmentation}

The semantic segmentation was obtained using RandLA-net \cite{hu2019randla}, with a pre-trained model provided by the authors. The model was trained using Semantic KITTI data set \cite{behley2019iccv} and therefore uses the labels from that set. The semantic classes contain separate labels for corresponding semi-static and dynamic objects, such as a car and a moving car, but the network could not reliably detect dynamic objects. The semantic segmentation results were noisy but sufficient to enable the experiments. One assumed main contributor to label noise is the domain transfer from one laser sensor to another, as the Semantic KITTI data set is recorded from a 64-channel laser. In contrast, the laser used in the Oxford Radar RobotCar data set has 32 channels. However, retraining the network with semi-synthetic measurements transformed using a domain transfer method \cite{langer2020iros} did not improve the segmentation accuracy.

\subsection{Filtering}
\label{sec:filters}

We use two filters to implement the function $f_d(z)$ for partitioning the measurements into the dynamic classes.

First, a dynamic filter removes measurements originating from dynamic objects. The filter removes the ground plane and clusters the remaining points. The cluster centroids are stored and associated with the cluster centroids of the subsequent measurement. The estimated movement of the cluster centroids combined with the semantic labels was used to determine whether the cluster represents a dynamic or non-dynamic object.

Second, a semantic filter removes all measurements with non-static semantic labels. We consider labels 40--99 from Semantic KITTI as static.

\subsection{Map creation}
\label{sec:map-creation}

Two maps were created from two sequences, yielding four maps.

The first map is the state-of-the-art baseline \gls{ndt-om}, created using all measurements, \ie{} $\delta_m = \{ \mathcal{S}, \mathcal{E}, \mathcal{D} \}$. However, as the measurement model of \gls{ndt-om} removes the dynamic objects, the map contains only static and semi-static objects. This method is referred to as the baseline mapping method.

The second map uses only static measurements, \ie{} $\delta_m = \{ \mathcal{S} \}$. This map is created using the semantic label filter (Section \ref{sec:filters}) to retain only measurements resulting from static objects and is referred to as the static mapping method.

Both maps were created using \gls{ndt-om} fusion method \cite{stoyanov_normal_2013} using ground truth poses, voxel size of $0.6$ m, inside the range proposed by the authors, and sub-maps with the dimensions $(x,y,z)$ of $(200, 200, 20)$m. All of the experiments were run at $0.2$ of real-time.

\begin{table}[t]
\caption{The used localization methods}
\tabcapspace{}
\begin{center}
\scalebox{0.9}{
\begin{tabular}{|c|c|c|c|}
\hline
\textbf{Name} & \textbf{ Dynamic filter } & \textbf{ Semantic filter } & \textbf{ $\delta_z$ }  \\
\hline
    baseline & -          & -          & $ \{ \mathcal{S}, \mathcal{E}, \mathcal{D}\} $ \\
    filtered & \checkmark & -          & $ \{ \mathcal{S}, \mathcal{E}\} $ \\
    static   & -          & \checkmark & $ \{ \mathcal{S} \} $ \\
    combined & \checkmark & \checkmark & $ \{ \mathcal{S} \} $ \\
\hline
\end{tabular}
}
\label{tab:weights}
\end{center}
\figvspace{}
\end{table}
\begin{table}[t]
\caption{The used motion model of \gls{ndt-mcl} }
\tabcapspace{}
\begin{center}
\begin{tabular}{|c|c|c|}
\hline
\textbf{Parameter} & \textbf{ Variance (t) } & \textbf{ Variance ($\psi$)} \\
\hline
    $x$         & 0.100 m    & 0.05 rad\\
    $y$         & 0.050 m    & 0.05 rad\\
    $z$         & 0.050 m    & 0.01 rad\\
    $\varphi$   & 0.010 m    & 0.01 rad\\
    $\theta$    & 0.010 m    & 0.01 rad\\
    $\psi$      & 0.001  m   & 0.05 rad\\
\hline
\end{tabular}
\label{tab:mcl-motionmodel}
\end{center}
\figvspace{}
\end{table}

\subsection{Localization}

To study the selection of $\delta_z$ presented in Section \ref{sec:methods}, we pre-process the measurements using the filters presented in Section \ref{sec:filters} and localize using \gls{ndt-mcl} \cite{saarinen_mcl_2013}, creating four localization methods: one with each filter, one without any filtering, and one with both filters. The methods, the dynamic content of the measurements, and the applied filtering methods are presented in Table \ref{tab:weights}. The baseline method uses all measurements, while the filtered method uses the semi-static and static measurements. The static and the combined methods use only the static measurements.

The parameterization of \gls{ndt-mcl} is the same in all methods and experiments. We use the same motion model in \cite{saarinen_mcl_2013}, with variances in Table \ref{tab:mcl-motionmodel}. As the robot moves in a planar environment in the experiments, the state is constrained to $x_t = [x, y, \psi]$, where $\psi$ is the yaw angle. Localization was initialized around the known initial pose $x_0$ with uniform distribution with dimensions $[-20, 20]$m in $x, y$ axes, $[0, 2\pi]$rad in $\psi$, and $\emptyset$ in axes $z, \varphi, \theta$. All of the experiments were run at $0.4$ of real-time.

\subsection{Metrics}

The estimated pose was stored at each time step, as well as the ground truth. Two metrics were calculated: \gls{rmse} of \gls{ate} and \gls{rpe}, which was calculated between two consecutive pose estimates \cite{sturm_benchmark_2012}.

\subsection{Results}

\begin{figure}[t]
\centering
\includegraphics[width=\linewidth]{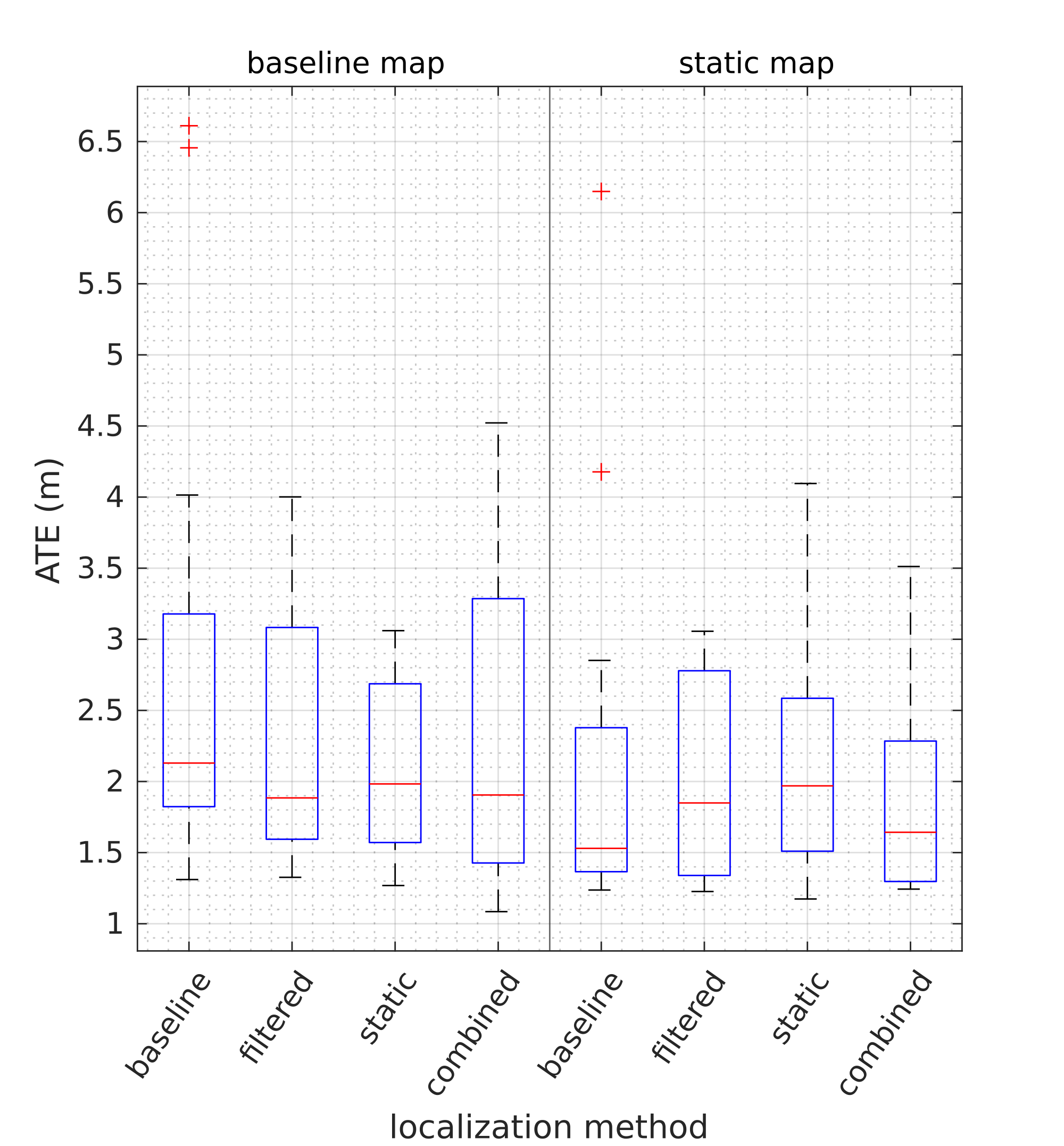}
\caption{The experiment results. In the figure, the sample median is presented with a red line, and the blue box represents the range between 25$^{\text{th}}$ and 75$^{\text{th}}$ percentile, \ie{} the interquartile range. The black dashed line presents the interval between the minimum and the maximum samples. Values over 1.5 times the interquartile range are marked as outliers, and displayed with a red plus symbol.
}
\label{fig:results}
\figvspace{}
\end{figure}

\begin{table*}[t!]
    \caption{Average results and variances over all experiments}
    \tabcapspace{}
    \begin{center}
        \setlength\tabcolsep{6pt}
        \scalebox{1.0}{
                \begin{tabular}{lllllll}
                    \toprule
                    & & \multicolumn{2}{c}{Mean} & & \multicolumn{2}{c}{Variance} \\
                    \textbf{Metric} & \textbf{Localization type} & \textbf{Baseline map} & \textbf{Static map} & & \textbf{Baseline map} & \textbf{Static map} \\
                    \midrule
                    \multirow{4}*{ATE}
                    & baseline & 2.8391 m  & 2.1629 m & & 2.9042 m  & 1.9628 m  \\
                    & filtered & 2.2503 m  & 1.9615 m & & 0.7282 m  & \textbf{0.4319 m} \\
                    & static & \textbf{2.1012 m}  & 2.1342 m & & \textbf{0.3881 m}  & 0.6058 m  \\
                    & combined & 2.2414 m  & \textbf{1.8749 m} & &  1.1353 m  & 0.4750 m  \\
                    \midrule
                    \multirow{4}*{RPE}
                    & baseline & 0.8958 m  & 0.8155 m  & & 0.8958 m  & 0.8155 m  \\
                    & filtered & 0.8571 m  & 0.7851 m  & & 0.8571 m  & 0.7851 m  \\
                    & static & 0.8877 m  & 0.8053 m  & & 0.8877 m  & 0.8053 m  \\
                    & combined & 0.8271 m  & 0.7697 m & & 0.8271 m  & 0.7697 m \\
                    \bottomrule
                \end{tabular}
        }
        \setlength\tabcolsep{6pt}
        \label{tab:results-table}
    \end{center}
    \figvspace{}
\end{table*}

Several conclusions can be drawn from the results in terms of \gls{ate}, which are presented in Figure \ref{fig:results}. The means of \gls{ate}s over all experiments are presented in Table \ref{tab:results-table}.

First, using the static map improves localization accuracy, which can be seen from Figure \ref{fig:results} by comparing the performance of each localization method over the two different types of map. With all methods except static localization, using the static map would be preferable as it reduces variance, improves the mean, or both. With static localization, the difference between the maps is negligible. This is likely due to the nature of NDT registration, where only matches between measurements and the map contribute to the cost. As there is no cost for unmatched cells, it matters less if the measurements are removed from the measurements or the map, as the reduction in error is similar.

The static map increases performance in three of the four cases, and in one case, the performance stays the same. As the static map consists of only static measurements, this result concurs with the hypothesis that having consistent assumptions over dynamics increases localization accuracy. This is illustrated in Figure \ref{fig:example}, where a parking lot is depicted, where the strengths of this method are most visible. Since the semi-static environment changes configuration rapidly in the form of different spaces being occupied and free each time, the semi-static measurements do not provide reliable landmarks. The static map omits these measurements and provides a reliable map against which to localize.

Second, the filtering of the measurements during localization also improves localization accuracy. Figure \ref{fig:results} suggests that dynamic objects may cause large magnitude errors when incorrectly matched with the map. Our experiments showed that filtering dynamic and semi-static objects decreased the maximum errors and removed the large outliers. Compared to the baseline localization, the filtering methods have reduced mean, variance, or both, making them more desirable choices.

Third, in terms of variance, static localization performs best. Whereas filtered localization can achieve very low errors, the variance is higher than static localization. While using more measurements is generally beneficial for localization accuracy, the incorrect matching of semi-static objects may cause errors. This makes using only static objects desirable, as they are the most reliable landmarks.

Given the two main hypotheses: (i) using only static measurements in the map and (ii) filtering the localization input are both beneficial for the localization accuracy, it should follow that the baseline localization with the baseline map should be on average the worst-performing combination, which can be seen from the results. As the baseline map holds semi-static measurements and the localization uses dynamic measurements, these can be incorrectly matched, reducing performance. Therefore, the localization accuracy is decreased by violating the consistent assumptions over dynamics. Conversely, when the static map and the combined method are used, the minimum \gls{ate} over all combinations is achieved.

The means and variances in terms of \gls{rpe} are presented in Table \ref{tab:results-table}. The differences between the different mapping and localization methods are negligible.

The effect of increased temporal distance between the map creation and localization time was studied, but the experiments were inconclusive. This is likely because the data set was gathered at a very similar time of day with respect to the traffic conditions, namely, from 11:46 to 15:20 during weekdays. The environment is likely more similar at the same time of day across different days than between different times on the same day. Therefore, a more heterogeneous data set should be acquired to study the temporal effects better.

In conclusion, the results indicate that localization accuracy increases when both semantic and dynamic information is considered.

\newcommand\imageScale{0.15}
\newcommand\divOne{0.245}

\begin{figure*}[t]

\begin{subfigure}[t]{\divOne\textwidth}
    \centering
    \scalebox{\imageScale}{\includegraphics{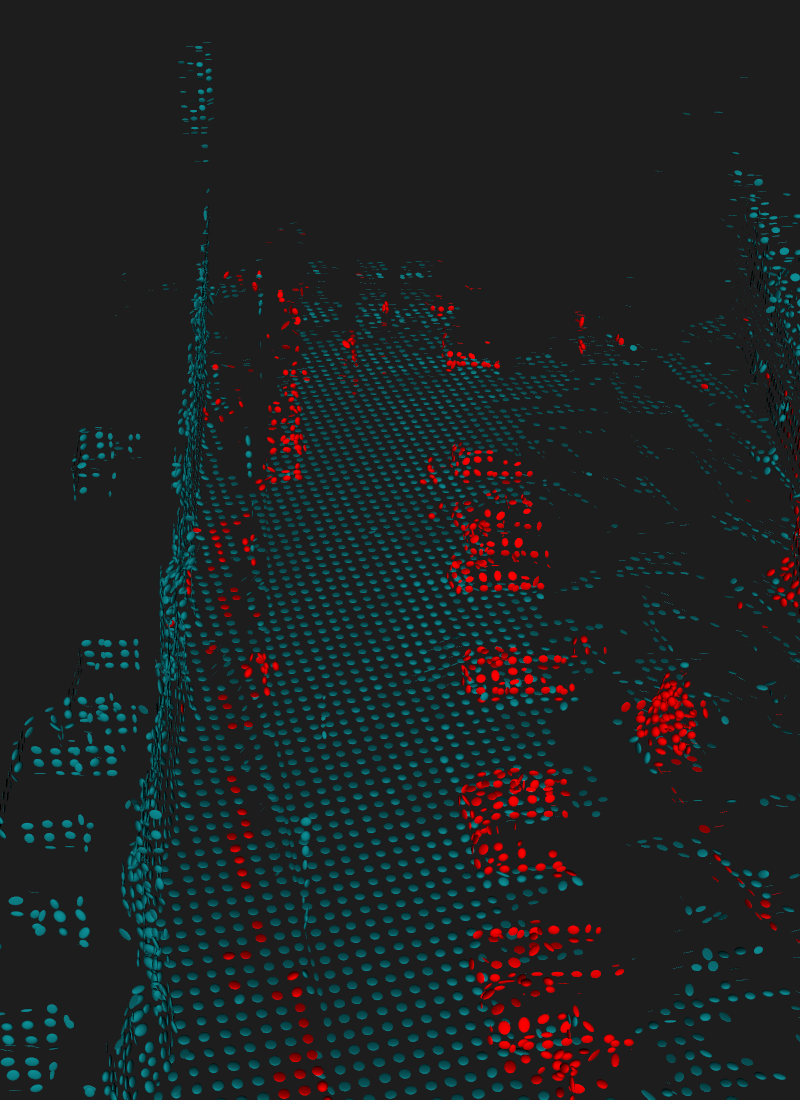}}
    \caption{}
    \label{fig:example1}
\end{subfigure}
\begin{subfigure}[t]{\divOne\textwidth}
    \centering
    \scalebox{\imageScale}{\includegraphics{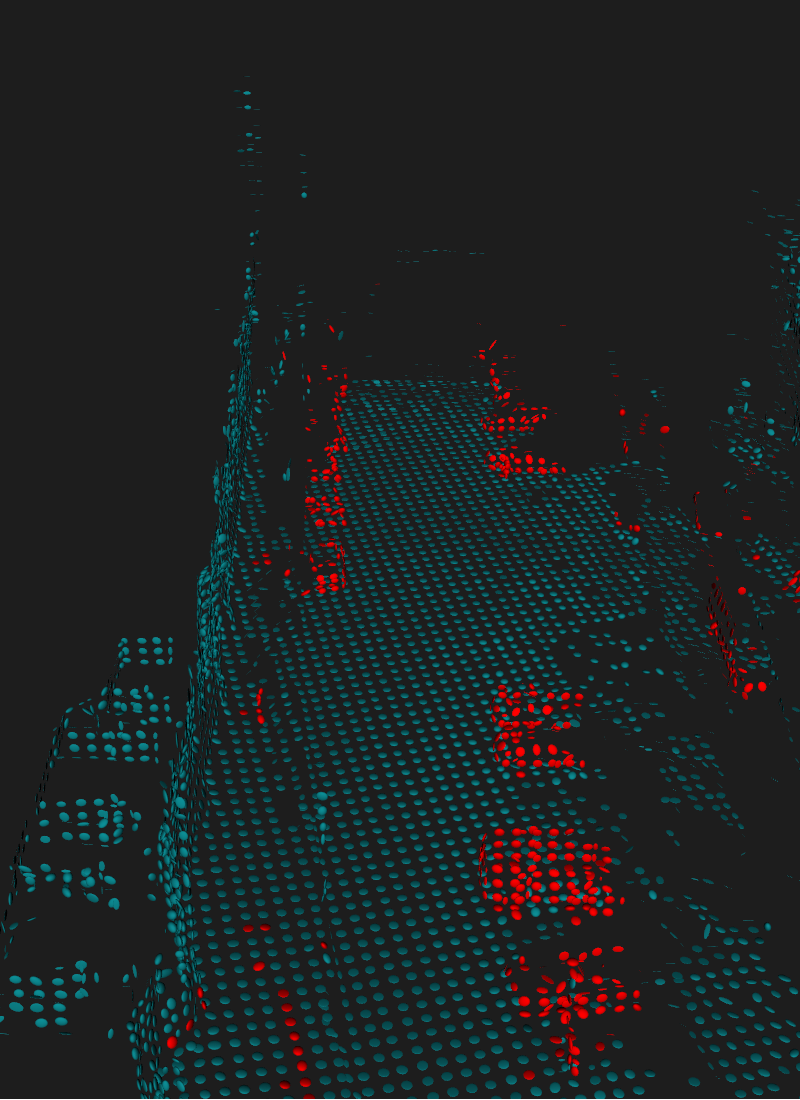}}
    \caption{}
    \label{fig:example2}
\end{subfigure}
\begin{subfigure}[t]{\divOne\textwidth}
    \centering
    \scalebox{\imageScale}{\includegraphics{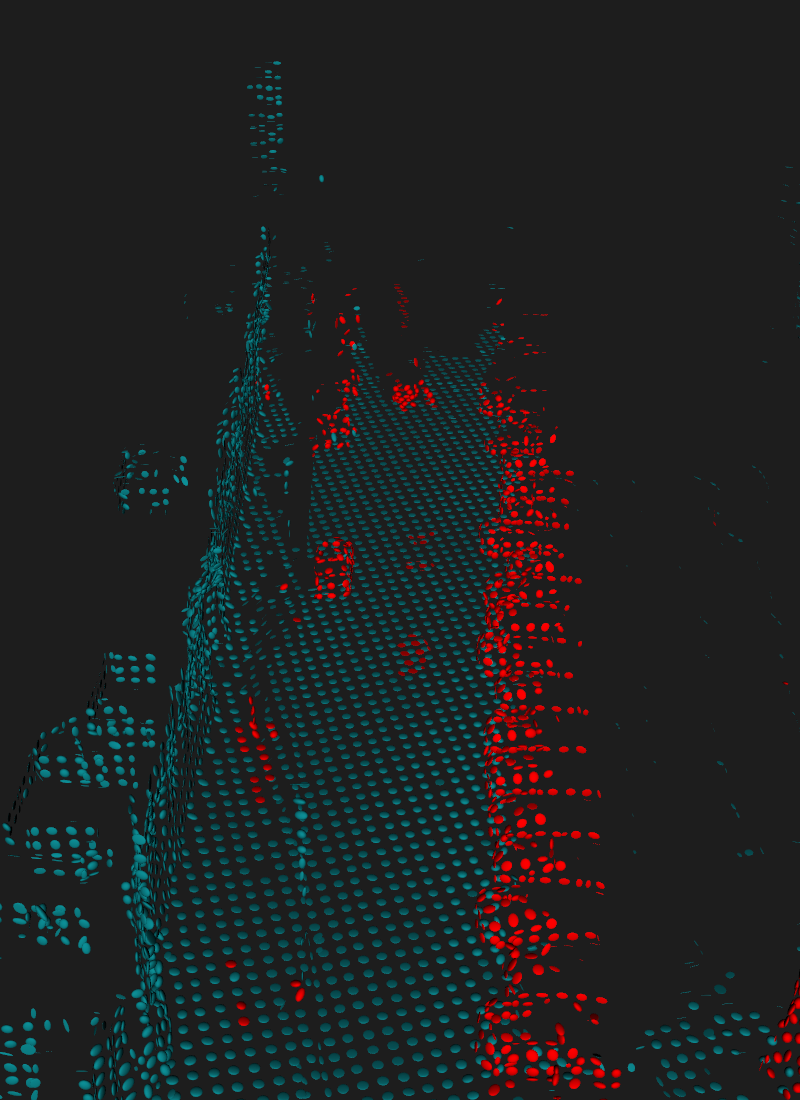}}
    \caption{}
    \label{fig:example3}
\end{subfigure}
\begin{subfigure}[t]{\divOne\textwidth}
    \centering
    \scalebox{\imageScale}{\includegraphics{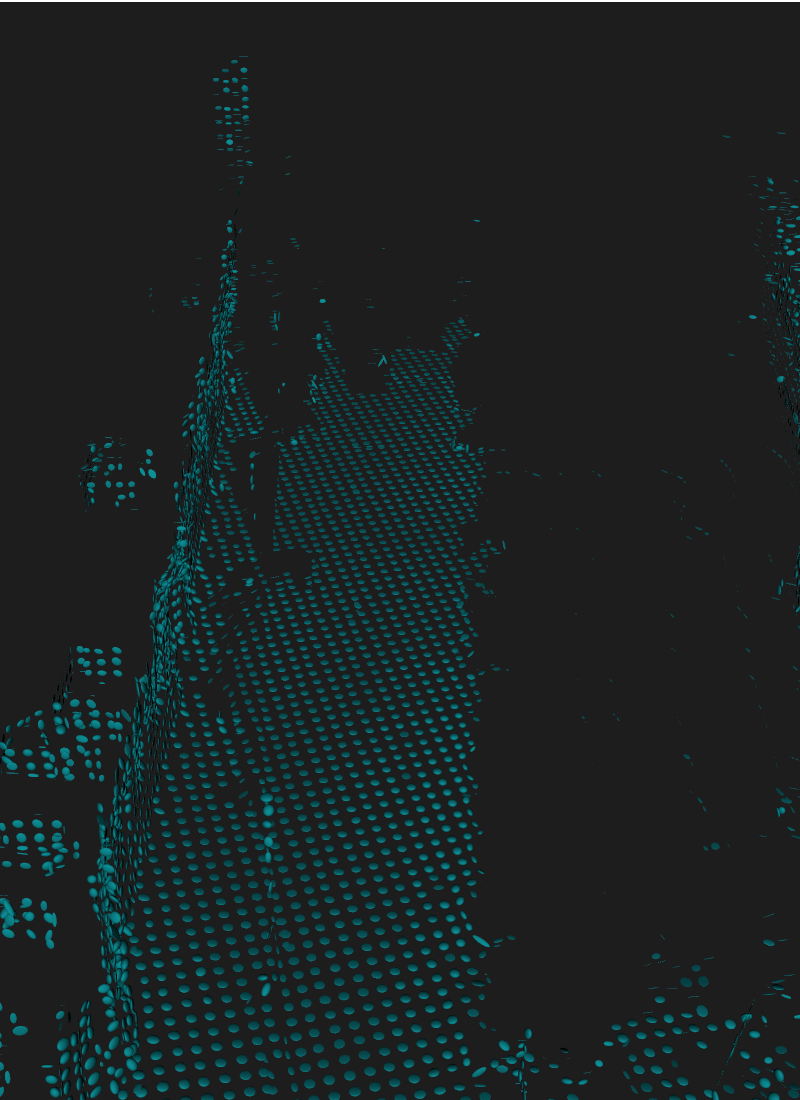}}
    \caption{}
    \label{fig:example4}
\end{subfigure}
\caption{A parking lot is a common scene where a large portion of the measurements originate from unreliable semi-static landmarks, shown in red in the baseline map. The configuration of the parking lot changes frequently, as seen from the maps created on different days (a)-(c). When the same measurements are filtered, only reliable static landmarks remain in the static map, as shown in (d).}
\label{fig:example}
\figvspacelast{}
\end{figure*}

\section{Conclusion}
\label{sec:conclusion}

In this work, we argue that the dichotomy of dividing the world into static and dynamic components is not sufficient for mapping. We showed that violating the static world assumption increases the localization error due to the mismatch between the map and semi-static or dynamic measurements treated as static. We additionally proposed a method to partition measurements according to their dynamic properties through dynamic object filtering and semantic segmentation. Finally, we used the proposed method to build a mapping-localization framework consistent with assumptions over dynamics.

The proposed methods were tested with 112 localization experiments with real-world data gathered over seven days spanning nine days in a city traffic scenario. The results show that by using a map consisting of static measurements, using only static measurements for localization, or both, the localization error lowered in terms of \gls{ate}. More importantly, the variance of the error decreased significantly. 

While the data set in this work was gathered in a relatively static urban setting, the proposed methods would likely be even more useful in environments containing more semi-static objects. However, in environments with few or no static objects, the proposed method must be extended to leverage the measurements from other dynamic classes without relaxing the consistent assumptions over dynamics.

In conclusion, we showed that consistent assumptions over dynamics increase localization accuracy. This result highlights the need for more realistic models of dynamics. The use of such models could enable localization in more challenging environments where current methods fail. Moreover, while in this work, we studied only localization accuracy, the proposed methods and assumptions could improve performance in other essential capabilities of mobile robotics, such as mapping and path planning.

\bibliographystyle{IEEEtran}
\bibliography{clean-abbr,ctrl}

\end{document}